\documentclass{article}
\usepackage{spconf,amsmath,graphicx,hyperref}
\usepackage{tabularx} 
\usepackage{array} 
\usepackage{amssymb}
\usepackage{multirow}
\usepackage{multicol}
\usepackage[dvipsnames]{xcolor}

\title{Class-Invariant Test-Time Augmentation for Domain Generalization}
\name{Zhicheng Lin$^{*}$, Xiaolin Wu$^{*}$, Xi Zhang$^{\dagger}$}
\address{$^{*}$School of Computing and Artificial Intelligence, Southwest Jiaotong University, China\\
$^{\dagger}$ANGEL Lab, Nanyang Technological University, Singapore\\
Github: \textcolor{magenta}{https://github.com/lzc0907/CI-TTA}}
\begin{document}
\ninept

\makeatletter
\everydisplay\expandafter{%
  \the\everydisplay
  \abovedisplayskip=4pt
  \belowdisplayskip=4pt
  \abovedisplayshortskip=4pt
  \belowdisplayshortskip=4pt
}
\makeatother

\maketitle
\begin{abstract}

Deep models often suffer significant performance degradation under distribution shifts. 
Domain generalization (DG) seeks to mitigate this challenge by enabling models to generalize to unseen domains. 
Most prior approaches rely on multi-domain training or computationally intensive test-time adaptation. 
In contrast, we propose a complementary strategy: lightweight test-time augmentation.  
Specifically, we develop a novel Class-Invariant Test-Time Augmentation (CI-TTA) technique.  The idea is to generate multiple variants of each input image through elastic and grid deformations that nevertheless belong to the same class as the original input. 
Their predictions are aggregated through a confidence-guided filtering scheme that remove unreliable outputs, 
ensuring the final decision relies on consistent and trustworthy cues.  
Extensive Experiments on PACS and Office-Home datasets demonstrate consistent gains across different DG algorithms and backbones, highlighting the effectiveness and generality of our approach.

\end{abstract}
\section{Introduction}
\label{sec:intro}

Although deep learning has achieved remarkable success in many controlled tasks~\cite{ResNet}, its performance often degrades significantly when confronted with new data from different domains. This highlights the vulnerability of models under distribution shifts. To address this challenge, domain generalization (DG)~\cite{wang2022generalizing} has emerged as a critical and challenging research area. DG aims to train models that acquire generalizable knowledge from one or more source domains and apply it to unseen target domains, without requiring access to target data. Such a capability is essential for deploying models in real-world scenarios where distribution shifts are ubiquitous.

Domain generalization has been extensively studied, with most efforts focusing on training-time strategies. These include employing larger models and more diverse datasets~\cite{orhan2019robustness}, as well as domain-invariant representation learning~\cite{ganin2016domain,coral}, meta-learning~\cite{mldg}, regularization techniques~\cite{vrex,groupdro,rsc}, and data augmentation~\cite{zhang2017mixup}. While effective to some extent, such approaches typically involve complex optimization procedures, cannot be adapted once deployed, and overlook test-time information that could provide valuable cues for improving model robustness.
To overcome these limitations, test-time adaptation (TTAda)\cite{liang2025comprehensive} has gained increasing attention. By exploiting test data, models can be adapted through partial weight updates\cite{iwasawa2021test,chen2023improved}, normalization statistic calibration~\cite{zhang2022memo}, feature-space style alignment~\cite{park2023test, zhao2022test}, or the use of prior knowledge. While effective, these TTAda methods often rely on heavy-duty backpropagation, source-specific pretraining, or auxiliary tasks, which increase inference latency and compromise their plug-and-play usability.

A lighter alternative is test-time augmentation (TTA), which aggregates predictions over simple transformations such as flips, crops, and color jitter~\cite{lyzhov2020greedy}. TTA is attractive because it requires no model modification. However, its utility in DG is limited: naive transformations do not explicitly address domain shifts, and uniform averaging across predictions may even degrade performance~\cite{tta}. This motivates the need for augmentation schemes that are not only lightweight but also domain-relevant.
Recent studies suggest that shape provides a particularly stable and domain-invariant cue. Standard CNNs are often biased toward texture, whereas enhancing shape bias improves robustness~\cite{geirhos2018imagenetshape1}. Deep networks trained on ImageNet have also shown a human-like preference for shape, which supports generalization~\cite{ritter2017cognitiveshape2}. Moreover, shape-based augmentation has been reported to improve generalization in medical image segmentation~\cite{castro2018elastic}. These findings highlight the fundamental role of shape information and inspire us to design class-invariant test-time augmentations for DG.

\begin{figure}[t]
  \centering
  \begin{minipage}[t]{0.3\linewidth}
    \centering
    \includegraphics[width=0.95\linewidth]{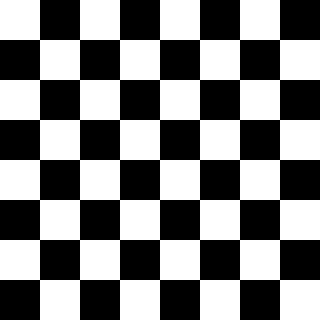}
  \end{minipage}
  \begin{minipage}[t]{0.3\linewidth}
    \centering
    \includegraphics[width=0.95\linewidth]{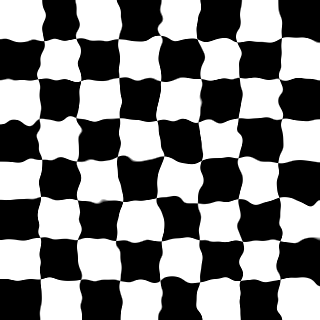}
  \end{minipage}
  \begin{minipage}[t]{0.3\linewidth}
    \centering
    \includegraphics[width=0.95\linewidth]{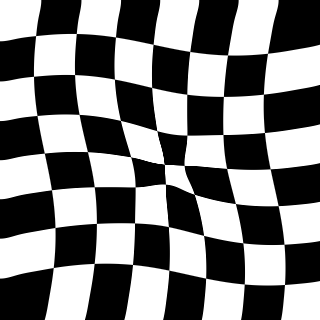}
  \end{minipage} \\
  \begin{minipage}[t]{0.3\linewidth}
    \centering
    \includegraphics[width=0.95\linewidth]{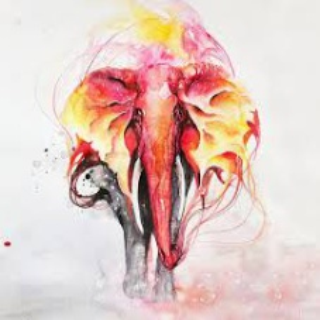}
    \vspace{2pt}
    {\footnotesize Original}
  \end{minipage}
  \begin{minipage}[t]{0.3\linewidth}
    \centering
    \includegraphics[width=0.95\linewidth]{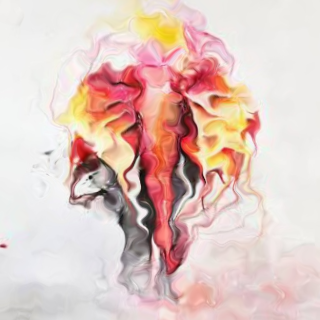}
    \vspace{2pt}
    {\footnotesize Elastic deformation}
  \end{minipage}
  \begin{minipage}[t]{0.3\linewidth}
    \centering
    \includegraphics[width=0.95\linewidth]{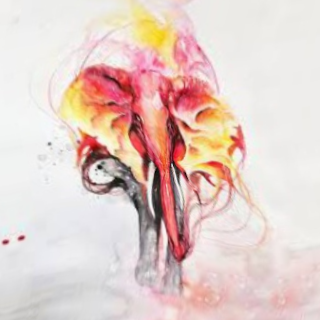}
    \vspace{2pt}
    {\footnotesize Grid deformation}
  \end{minipage}
\vspace{-6pt}
\caption{Examples of class-invariant test-time augmentation.}
  \label{fig:pacs_sp_tta}
\vspace{-12pt}
\end{figure}

Building on these insights, we propose Class-Invariant Test-Time Augmentation (CI-TTA), a plug-and-play strategy that improves domain generalization at inference time. CI-TTA introduces elastic and grid deformations (see Fig.~\ref{fig:pacs_sp_tta}) that generate diverse, class-consistent variants of test inputs, encouraging models to exploit domain-invariant shape cues rather than textures. To further enhance reliability, we design a confidence-guided aggregation scheme that filters out low-confidence predictions, ensuring that final decisions rely on stable and trustworthy cues, in line with the principle of prediction self-consistency~\cite{nguyen2020ensemble,wang2022self}. Beyond methodological novelty, we provide extensive empirical validation on PACS~\cite{pacs} and Office-Home~\cite{office-home} datasets, demonstrating consistent gains across diverse DG algorithms and backbones. Together, these contributions establish CI-TTA as a lightweight, effective, and broadly applicable approach for improving DG robustness under distribution shifts.

\vspace{-1em} 
\begin{figure*}[!ht]
  \centering
  \includegraphics[width=0.96\textwidth]{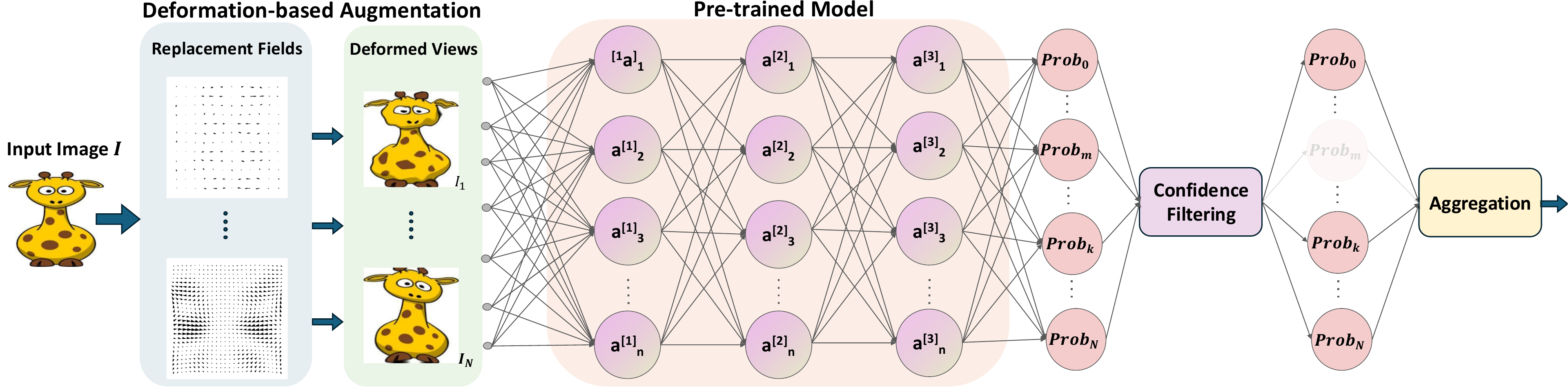}
  \caption{Overview of the proposed CI-TTA framework for efficient domain generalization.}
  \label{fig:framework}
\end{figure*}
\vspace{-1em}

\section{Method}
\label{sec:methods}
The overall framework of our proposed Class-Invariant Test-Time Augmentation (CI-TTA) is illustrated in Fig.~\ref{fig:framework}.
Given a pre-trained model $M$ and a test image $I \in \mathbb{R}^{H \times W \times C}$, the inference process consists of four sequential stages: augmentation, prediction, filtering, and aggregation.
We describe each stage in detail below.

\textbf{Deformation-based Augmentation.}
Let $\mathcal{T}={T_1, \dots, T_N}$ denote a set of stochastic deformation operators.
Applying them to $I$ yields a collection of augmented test images
\begin{equation}
\mathcal{I}' = \{I_i = T_i(I)\}_{i=1}^N.
\end{equation}
Each $T_i$ is instantiated from either an elastic deformation or a grid deformation, both designed to preserve global shape while introducing local variations.
To ensure controlled randomness, deformation parameters are sampled from a zero-mean Gaussian distribution
\begin{equation}
\theta_i \sim \mathcal{N}(0, \sigma^2),
\end{equation}
where $\sigma$ controls the perturbation magnitude.
This guarantees that displacements remain unbiased (expected zero shift), thereby avoiding systematic translation or distortion of object structures.

\textit{Elastic Deformation.}  
We define a random displacement field $\mathbf{d}(x) \in \mathbb{R}^2$ for each pixel coordinate $x \in \Omega \subset \mathbb{R}^2$.
First, a noise field $\mathbf{z}(x) \sim \mathcal{N}(0, \sigma^2)$ is generated.
A Gaussian smoothing kernel $G_\kappa$ (with kernel size $\kappa$) is then applied:
\begin{equation}
\mathbf{d}(x) = (G_\kappa * \mathbf{z})(x),
\end{equation}
where $*$ denotes convolution.
The deformation is realized by warping the original image $I$ with the displacement field $\mathbf{d}(x)$, yielding:
\begin{equation}
\setlength\abovedisplayskip{4pt}
\setlength\belowdisplayskip{4pt}
I'(x) = I\big(x + \mathbf{d}(x)\big).
\end{equation}
Here the Gaussian smoothing enforces spatial coherence, ensuring that perturbations are smooth and class-consistent.

\textit{Grid Deformation.}  
Let $\mathcal{P}={p_{ij}}$ denote a regular grid of control points over $\Omega$.
Each control point is perturbed by a random displacement $\delta_{ij} \sim \mathcal{N}(0, \sigma^2)$.
The new position is $ \tilde{p}{ij} = p{ij} + \delta_{ij} $.
An interpolation function $\Phi(\cdot)$, e.g., bicubic interpolation, maps pixel coordinates to the displaced grid:
\begin{equation}
\setlength\abovedisplayskip{4pt}
\setlength\belowdisplayskip{4pt}
I'(x) = I\big(\Phi(x; {\tilde{p}_{ij}})\big).
\end{equation}
By interpolating over the displaced control grid, the image undergoes smooth warping that introduces localized variations while maintaining global structural consistency.


With both aforementioned deformations unified under the Gaussian parameterization, CI-TTA achieves two benefits:
(1) comparability, since elastic and grid perturbations share a consistent scale controlled by $\sigma$, and
(2) simplicity, as hyperparameter tuning is reduced to a single variance parameter.


\textbf{Prediction from Original and Augmented Views.}
Given the original test image $I$ and the corresponding CI-TTA augmented variants $\mathcal{I}' = \{I_i\}_{i=1}^N$, we evaluate each input (including $I$ and $\mathcal{I}'$) using the pre-trained model $M:\mathbb{R}^{H\times W\times C}\rightarrow\mathbb{R}^K$, which maps an image of size $H \times W \times C$ to a $K$-dimensional logit vector corresponding to the class scores.
The resulting logit vectors are
\begin{equation}
\setlength\abovedisplayskip{3pt}
\setlength\belowdisplayskip{3pt}
\mathbf{z}_0 = M(I), \quad
\mathbf{z}_i = M(I_i), i=1,\dots,N.
\end{equation}
Each logit $\mathbf{z}_j \in \mathbb{R}^K$ is converted into a categorical probability distribution via the softmax operator:
\begin{equation}
\setlength\abovedisplayskip{3pt}
\setlength\belowdisplayskip{3pt}
\mathbf{p}_j = \text{softmax}(\mathbf{z}_j),
\quad j = 0,1,\dots,N.
\end{equation}
To quantify the reliability of each prediction, we define the confidence score as the maximum class probability:
\begin{equation}
\setlength\abovedisplayskip{3pt}
\setlength\belowdisplayskip{3pt}
C_j = \max_{k} [\mathbf{p}_j]_k,
\end{equation}
where $[\mathbf{p}_j]_k$ denotes the predicted probability of class $k$.
These confidence scores will be subsequently used to guide the filtering and aggregation process.






\textbf{Confidence-Guided Filtering.}
Not all augmented predictions are equally reliable, as severe deformations may introduce artifacts that mislead the model.
To suppress the influence of such noisy predictions, we incorporate a confidence-based filtering mechanism.
For each prediction $\mathbf{p}j$, its confidence score $C_j$ is compared against a fixed threshold $\tau \in [0,1]$.
Only distributions exceeding this threshold are retained:
\begin{equation}
\mathcal{P}_{\mathrm{f}} = \{ \mathbf{p}_j \mid C_j \geq \tau \}.
\end{equation}

This operation yields a refined set $\mathcal{P}_{\mathrm{f}}$ that excludes low-confidence predictions, ensuring that the subsequent aggregation relies solely on high-quality evidence.
In this way, the method improves robustness by preventing unreliable augmentations from diluting or biasing the final decision.








\textbf{Aggregation Strategy.}  
After filtering, let $\mathcal{P}_{\mathrm{f}} \subseteq \{\mathbf{p}_j\}_{j=0}^N$ denote the set of retained probability distributions.  
The final prediction distribution is obtained by averaging over $\mathcal{P}_{\mathrm{f}}$:  
\begin{equation}
\mathbf{p}_{*} 
= \frac{1}{|\mathcal{P}_{\mathrm{f}}|} 
\sum_{\mathbf{p}_j \in \mathcal{P}_{\mathrm{f}}} \mathbf{p}_j.
\end{equation}  

The corresponding class decision is then given by  
\begin{equation}
\hat{y}_* 
= \arg\max_{c} \; [\mathbf{p}_*]_c, \quad c \in \{1,\dots,K\}.
\end{equation}
where $[\mathbf{p}_{*}]_c$ denotes the probability assigned to class $c$.  

To guarantee a valid output in all circumstances, if $\mathcal{P}_{\mathrm{f}} = \varnothing$,  
the prediction is defaulted to the distribution $\mathbf{p}_0$ obtained from the original test image $I$.  
This ensures robustness of the framework even in the absence of reliable augmented predictions.

\section{Experiments and Results}
\label{sec:pagestyle}


This section begins by describing the experimental setup, followed by performance evaluation on standard DG benchmarks. We then conduct ablation studies to examine the effect of deformation strength, the contribution of confidence filtering, and the influence of the confidence threshold.



\subsection{Experimental Setting}
We conduct evaluations on the PACS~\cite{pacs} and Office-Home~\cite{office-home} benchmarks under the standard leave-one-domain-out DG protocol.  
Two backbones, ResNet-18~\cite{ResNet} and ResNet-50~\cite{ResNet}, are considered, and comparisons are made against representative DG algorithms spanning common methodological families: ERM~\cite{vapnik1998statistical}, DANN~\cite{ganin2016domain}, RSC~\cite{rsc}, VREx~\cite{vrex}, Mixup~\cite{zhang2017mixup}, and CORAL~\cite{coral}.  
All methods adopt a uniform training pipeline: input images are resized to $224{\times}224$ and normalized using the ImageNet mean and standard deviation.  
Optimization is performed with stochastic gradient descent (SGD) with momentum $0.9$ and weight decay $5{\times}10^{-4}$, using a fixed learning rate of $1{\times}10^{-3}$, batch size of $32$, and training for $120$ epochs.  
Training is performed on a single NVIDIA RTX 3090 GPU (24\,GB) under the same software environment (PyTorch 2.2.0 / CUDA 11.8 on Linux), ensuring fairness across algorithms and backbones.  

For each input, $N{=}100$ class-invariant variants are generated using a $4{\times}4$ grid deformation and an elastic deformation parameterized by the strength $\sigma$, in addition to the original image.  
Each view is evaluated to produce a softmax probability distribution; the confidence score is defined as the maximum predicted class probability.  
Only predictions with confidence not less than $\tau$ are retained for aggregation, with the prediction from the original image $I$ adopted as a fallback when no augmented views satisfy the threshold.  
The hyperparameters $\sigma$ and $\tau$ are selected based on performance on a held-out validation split (or following the settings reported in Sec.~\ref{sec:ablation}) and are fixed thereafter when reporting benchmark results.

\subsection{Performance on Domain Generalization Benchmarks}  
\label{sec:benchmarks}
We evaluate CI-TTA on both the PACS and Office-Home benchmarks with ResNet-18/50 backbones, applying the near-optimal $(\sigma,\tau)$ values selected for each model–dataset pair (see Sec.~\ref{sec:ablation} for sensitivity analysis).  
Top-1 accuracy is reported for each held-out target domain, along with the average accuracy across domains.  
The results are presented in Tables~\ref{tab:pacs-results} and \ref{tab:office-results}.  
Across all datasets, algorithms, and backbones, CI-TTA consistently improves generalization performance.  
On the PACS dataset, for instance, applying CI-TTA to ERM with ResNet-18 raises the mean accuracy from $80.56\%$ to $82.01\%$.  
Comparable improvements are observed for other algorithms such as DANN, RSC, and Mixup.  
On the Office-Home dataset, CI-TTA likewise yields consistent gains across baselines, further demonstrating its effectiveness and robustness.

\begin{table}[!t]
    \centering
    \vspace{-1em}
    \caption{Results of DG methods on PACS dataset with ResNet-18 and ResNet-50 backbones.
    We report average accuracy (\%) across domains. 
    “CI-TTA” denotes applying our method at test time.}
    \label{tab:pacs-results}
    \resizebox{0.95\linewidth}{!}{
    \begin{tabular}{c|c|cccc|c}
    \hline
     & Method & A & C & P & S & Avg \\
    \hline
    \hline
    \multirow{12}{*}{\rotatebox{90}{ResNet-18}} 
      & ERM~\cite{vapnik1998statistical}   & 79.44 & 75.38 & 96.35 & 71.06 & 80.56 \\
      & ERM + CI-TTA & \textbf{80.22} & \textbf{75.81} & \textbf{96.57} & \textbf{75.45} & \textbf{82.01} \\
      \cline{2-7}
      & DANN~\cite{ganin2016domain} & 81.79 & 75.04 & 95.39 & 74.75 & 81.74 \\
      & DANN + CI-TTA & \textbf{82.27} & \textbf{76.11} & \textbf{95.51} & \textbf{77.32} & \textbf{82.80} \\
      \cline{2-7}
      & RSC~\cite{rsc}   & 80.57 & 75.21 & 96.29 & 74.27 & 81.59 \\
      & RSC + CI-TTA & \textbf{81.78} & \textbf{75.40} & \textbf{96.38} & \textbf{77.27} & \textbf{82.70} \\
      \cline{2-7}
      & VREx~\cite{vrex}  & 80.03 & 75.47 & 96.53 & 72.38 & 81.10 \\
      & VREx + CI-TTA & \textbf{80.66} & \textbf{75.76} & \textbf{96.56} & \textbf{75.64} & \textbf{82.15} \\
      \cline{2-7}
      & Mixup~\cite{zhang2017mixup} & 81.69 & 74.45 & 95.21 & 76.10 & 81.86 \\
      & Mixup + CI-TTA & \textbf{83.39} & \textbf{75.49} & \textbf{95.65} & \textbf{77.87} & \textbf{83.25} \\
      \cline{2-7}
      & CORAL~\cite{coral} & 79.35 & 74.19 & 93.17 & 81.11 & 81.95 \\
      & CORAL + CI-TTA & \textbf{80.17} & \textbf{74.83} & \textbf{93.29} & \textbf{82.82} & \textbf{82.77} \\
    \hline
    \hline
    \multirow{12}{*}{\rotatebox{90}{ResNet-50}} 
      & ERM~\cite{vapnik1998statistical}   & 85.64 & 79.27 & 97.10 & 79.33 & 85.34 \\
      & ERM + CI-TTA & \textbf{86.77} & \textbf{79.33} & \textbf{98.00} & \textbf{80.43} & \textbf{86.13} \\
      \cline{2-7}
      & DANN~\cite{ganin2016domain} & 85.64 & 79.96 & 97.54 & 80.40 & 85.89 \\
      & DANN + CI-TTA & \textbf{86.23} & \textbf{80.13} & \textbf{98.08} & \textbf{80.49} & \textbf{86.23} \\
      \cline{2-7}
      & RSC~\cite{rsc}   & 86.43 & 79.01 & 97.66 & 79.41 & 85.63 \\
      & RSC + CI-TTA & \textbf{87.79} & \textbf{79.35} & \textbf{98.14} & \textbf{79.51} & \textbf{86.20} \\
      \cline{2-7}
      & VREx~\cite{vrex}  & 85.84 & 78.80 & 97.20 & 79.49 & 85.33 \\
      & VREx + CI-TTA & \textbf{86.72} & \textbf{78.97} & \textbf{98.04} & \textbf{79.74} & \textbf{85.87} \\
      \cline{2-7}
      & Mixup~\cite{zhang2017mixup} & 87.67 & 82.00 & 96.59 & 79.26 & 86.38 \\
      & Mixup + CI-TTA & \textbf{87.96} & \textbf{82.05} & \textbf{97.13} & \textbf{81.12} & \textbf{87.07} \\
      \cline{2-7}
      & CORAL~\cite{coral} & 85.60 & 81.10 & 96.95 & 81.37 & 85.26 \\
      & CORAL + CI-TTA & \textbf{85.96} & \textbf{81.42} & \textbf{97.03} & \textbf{82.51} & \textbf{86.73} \\
    \hline
    \end{tabular}
    }
\end{table}

\begin{table}[!t]
    \centering
    \vspace{-1em}
    \caption{Results of DG methods on Office-Home dataset with ResNet-18/50 backbones.
    We report average accuracy (\%) across domains. 
    “CI-TTA” denotes applying our method at test time.}
    \label{tab:office-results}
    \resizebox{0.95\linewidth}{!}{
    \begin{tabular}{c|c|cccc|c}
    \hline
     & Method & A & C & P & R & Avg \\
    \hline
    \hline
    \multirow{12}{*}{\rotatebox{90}{ResNet-18}} 
      & ERM~\cite{vapnik1998statistical}   & 58.14 & 48.98 & 71.68 & 74.89 & 63.41 \\
      & ERM + CI-TTA & \textbf{58.43} & \textbf{49.42} & \textbf{71.82} & \textbf{74.92} & \textbf{63.64} \\
      \cline{2-7}
      & DANN~\cite{ganin2016domain}  & 57.44 & 47.88 & 71.21 & 73.42 & 62.49 \\
      & DANN + CI-TTA & \textbf{58.06} & \textbf{47.92} & \textbf{71.64} & \textbf{73.72} & \textbf{62.83} \\
      \cline{2-7}
      & RSC~\cite{rsc}   & 58.38 & 49.05 & 71.71 & 75.19 & 63.58 \\
      & RSC + CI-TTA & \textbf{58.47} & \textbf{49.16} & \textbf{72.01} & \textbf{75.42} & \textbf{63.77} \\
      \cline{2-7}
      & VREx~\cite{vrex}  & 58.38 & 49.03 & 71.66 & 75.07 & 63.54 \\
      & VREx + CI-TTA & \textbf{58.43} & \textbf{49.09} & \textbf{71.77} & \textbf{75.25} & \textbf{63.64} \\
      \cline{2-7}
      & Mixup~\cite{zhang2017mixup} & 58.22 & 49.76 & 71.75 & 74.45 & 63.55 \\
      & Mixup + CI-TTA & \textbf{59.66} & \textbf{49.84} & \textbf{72.49} & \textbf{74.48} & \textbf{64.12} \\
      \cline{2-7}
      & CORAL~\cite{coral} & 58.80 & 49.55 & 71.98 & 74.87 & 63.80 \\
      & CORAL + CI-TTA & \textbf{58.89} & \textbf{49.68} & \textbf{72.55} & \textbf{75.17} & \textbf{64.07} \\
    \hline
    \hline
    \multirow{12}{*}{\rotatebox{90}{ResNet-50}} 
      & ERM~\cite{vapnik1998statistical}   & 66.46 & 55.37 & 78.01 & 80.01 & 69.96 \\
      & ERM + CI-TTA & \textbf{66.53} & \textbf{55.40} & \textbf{78.53} & \textbf{80.13} & \textbf{70.14} \\
      \cline{2-7}
      & DANN~\cite{ganin2016domain}  & 65.22 & 54.45 & 78.53 & 79.02 & 69.31 \\
      & DANN + CI-TTA & \textbf{65.84} & \textbf{54.80} & \textbf{78.57} & \textbf{80.04} & \textbf{69.81} \\
      \cline{2-7}
      & RSC~\cite{rsc}   & 66.54 & 53.91 & 77.72 & 78.80 & 69.24 \\
      & RSC + CI-TTA & \textbf{67.12} & \textbf{55.26} & \textbf{77.74} & \textbf{79.97} & \textbf{70.02} \\
      \cline{2-7}
      & VREx~\cite{vrex}  & 66.63 & 55.51 & 78.76 & 80.33 & 70.31 \\
      & VREx + CI-TTA & \textbf{66.68} & \textbf{56.10} & \textbf{78.83} & \textbf{80.38} & \textbf{70.50} \\
      \cline{2-7}
      & Mixup~\cite{zhang2017mixup} & 67.74 & 56.32 & 77.83 & 79.66 & 70.39 \\
      & Mixup + CI-TTA & \textbf{68.36} & \textbf{57.94} & \textbf{77.92} & \textbf{79.68} & \textbf{71.01} \\ 
      \cline{2-7}
      & CORAL~\cite{coral} & 66.87 & 56.49 & 78.26 & 79.87 & 70.37 \\
      & CORAL + CI-TTA & \textbf{67.87} & \textbf{56.84} & \textbf{78.51} & \textbf{80.17} & \textbf{70.85} \\
    \hline
    \end{tabular}
    }
    \vspace{-1em}
\end{table}

\subsection{Ablation Studies}
\label{sec:ablation}
We conduct ablation studies to assess the contributions of three different components in CI-TTA: (i) the effect of deformation strength $\sigma$, (ii) the impact of incorporating or removing confidence filtering, and (iii) the influence of the confidence threshold $\tau$.

\begin{figure}[t]
\centering
\includegraphics[width=0.95\linewidth]{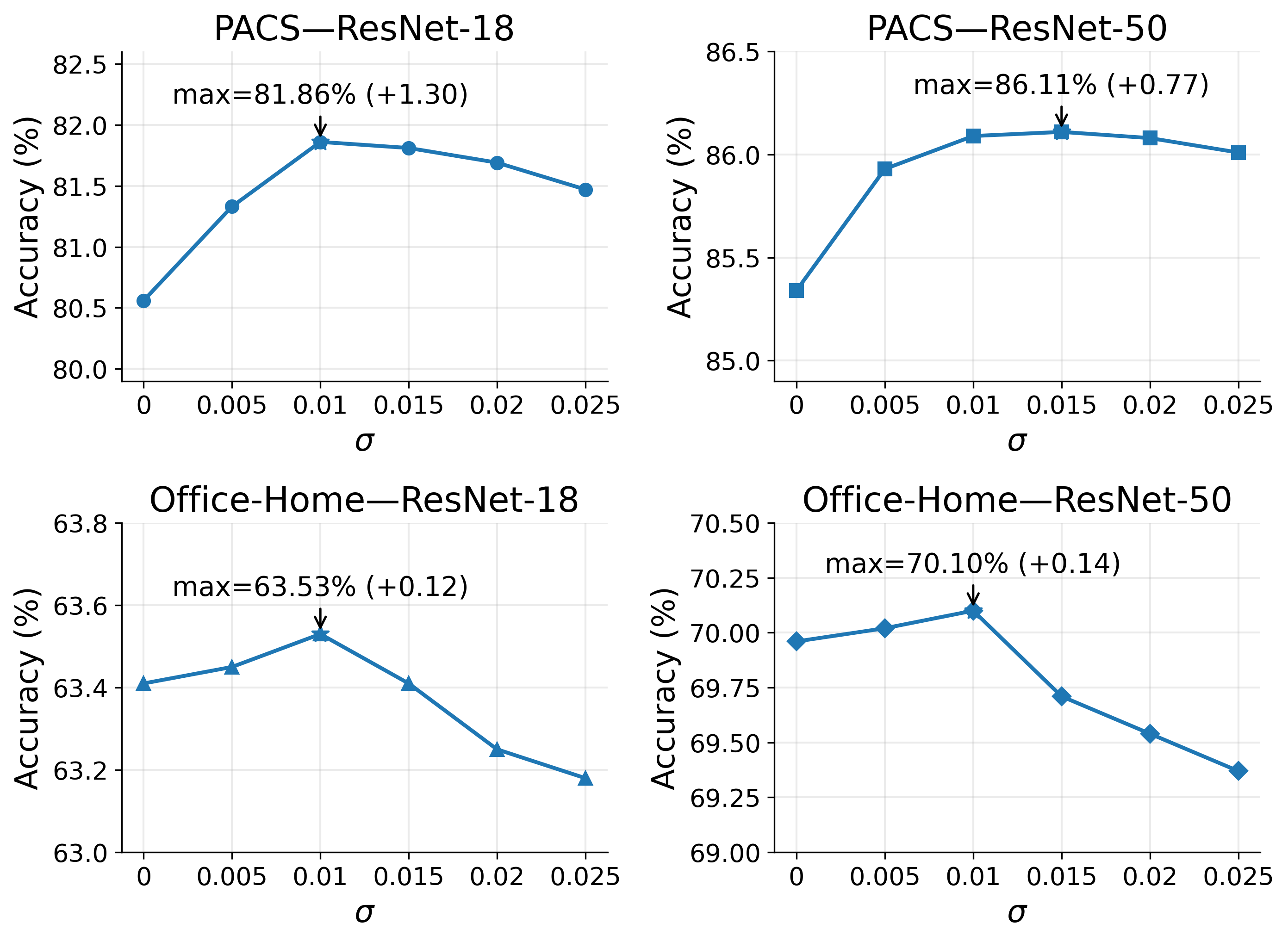}
\vspace{-1em}
\caption{Effect of deformation strength $\sigma$ in CI-TTA on accuracy under ERM training without confidence filtering. 
Left: PACS (ResNet-18/50). Right: Office-Home (ResNet-18/50).}

\label{fig:sigma_curve}
\end{figure}

\textbf{Impact of Deformation Parameters.}
As described in Sec.~\ref{sec:methods}, the deformation intensity is controlled by the standard deviation $\sigma$ of a zero-mean Gaussian distribution. To isolate its effect, we apply CI-TTA \emph{without} the confidence-based filtering module and vary $\sigma$ across a range of values on PACS and Office-Home with ResNet-18 and ResNet-50 backbones. For each $\sigma$, predictions are aggregated over $100$ deformed variants per test image. The case $\sigma=0$ corresponds to the baseline without deformation.
The empirical results (Fig.~\ref{fig:sigma_curve}) reveal consistent bell-shaped trends across all settings. On PACS, accuracy steadily increases from the baseline ($80.56\%$ with ResNet-18 and $85.34\%$ with ResNet-50) to peak performance near $\sigma \approx 0.01$–$0.015$, reaching $81.86\%$ and $86.11\%$, respectively. These gains, though modest in absolute value ($+1.3\%$ for ResNet-18 and $+0.8\%$ for ResNet-50), indicate that moderate perturbations can enhance robustness. On Office-Home, where the overall accuracies are lower, the effect is subtler: ResNet-18 improves from $63.41\%$ to $63.53\%$, while ResNet-50 rises from $69.96\%$ to $70.10\%$ around $\sigma=0.01$. 
Notably, when $\sigma$ becomes larger, accuracies decline across all models, confirming that excessive deformation introduces harmful distortions.
In summary, the sensitivity analysis demonstrates that CI-TTA benefits most from intermediate $\sigma$ values. Small perturbations ($\sigma < 0.005$) are too weak to bring measurable gains, while large perturbations ($\sigma > 0.02$) degrade performance. The consistent improvements across datasets and architectures highlight that controlled, class-invariant deformations provide useful diversity at test time, thereby improving under domain shifts.

    
    

    
    


\textbf{The Role of Confidence Filtering}
To assess the contribution of the confidence-based filtering mechanism,  
we conduct an ablation study across multiple DG algorithms.  
For each algorithm, two variants are compared:  
(1) \textbf{CI-TTA w/o CF}, which applies deformation-based augmentation and averages predictions over all augmented samples without filtering; and  
(2) \textbf{CI-TTA w/ CF}, the complete method incorporating both deformation and confidence-guided filtering.  
Table~\ref{tab:ablation} summarizes the results on PACS with a ResNet-18 backbone.  
The results indicate a consistent performance gap across all evaluated algorithms.  
Without filtering, direct averaging of augmented predictions yields only limited improvement.  
In contrast, the full CI-TTA achieves higher accuracy by suppressing unreliable predictions and aggregating only those with sufficient confidence.  
These findings highlight confidence-based filtering as a critical and generally applicable component,  
establishing CI-TTA as a robust inference-time strategy that can be seamlessly integrated with diverse DG algorithms to enhance generalization on unseen domains.



\begin{table}[t]
    \centering
    \vspace{-1em}
    \caption{Effect of confidence filtering in CI-TTA, evaluated on the PACS dataset with ResNet-18. 
    The variant with confidence filtering (“CI-TTA w/ CF”) achieves consistently higher accuracy 
    than the variant without filtering (“CI-TTA w/o CF”).}
    \label{tab:ablation}
    \begin{tabular}{c|c|c}
        \hline
        Algorithm & CI-TTA w/o CF & CI-TTA w/ CF \\
        \hline
        ERM   & 81.86 & \textbf{82.01} \\
        DANN~~\cite{ganin2016domain}  & 81.83 & \textbf{82.80} \\
        RSC~~\cite{rsc}   & 82.08 & \textbf{82.70} \\
        VREx~~\cite{vrex}  & 81.21 & \textbf{82.15} \\
        Mixup~~\cite{zhang2017mixup}   & 82.36 & \textbf{83.25} \\
        CORAL~~\cite{coral} & 82.16 & \textbf{82.77} \\
        \hline
    \end{tabular}
    \vspace{-1em}
\end{table}

\begin{figure}[t]
\centering

\begin{minipage}[b]{0.245\linewidth}
  \centering
  \centerline{\tiny{PACS, ResNet-18}}
  \includegraphics[width=\linewidth]{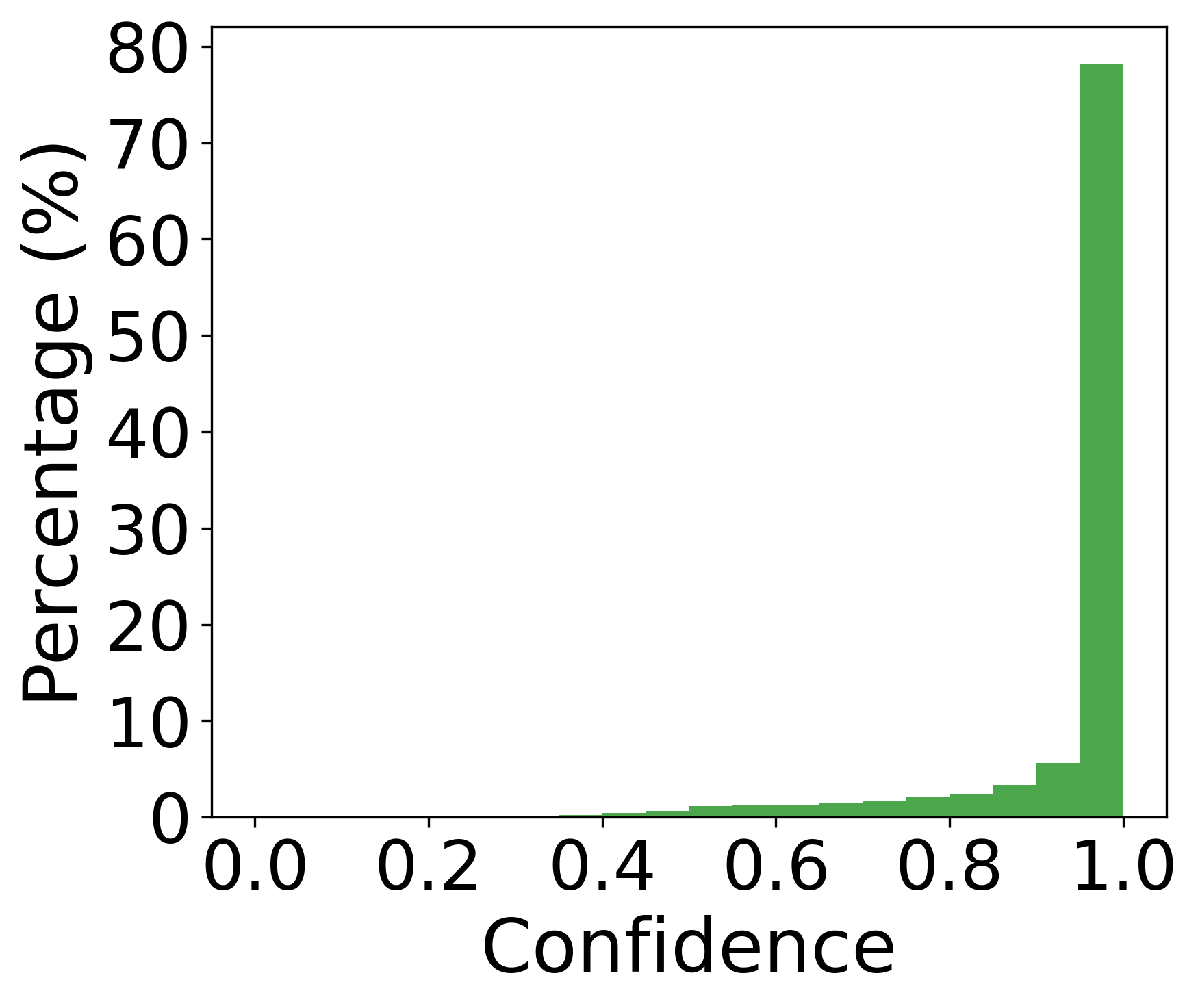}
\end{minipage}\hfill
\begin{minipage}[b]{0.245\linewidth}
  \centering
  \centerline{\tiny{PACS, ResNet-18}}
  \includegraphics[width=\linewidth]{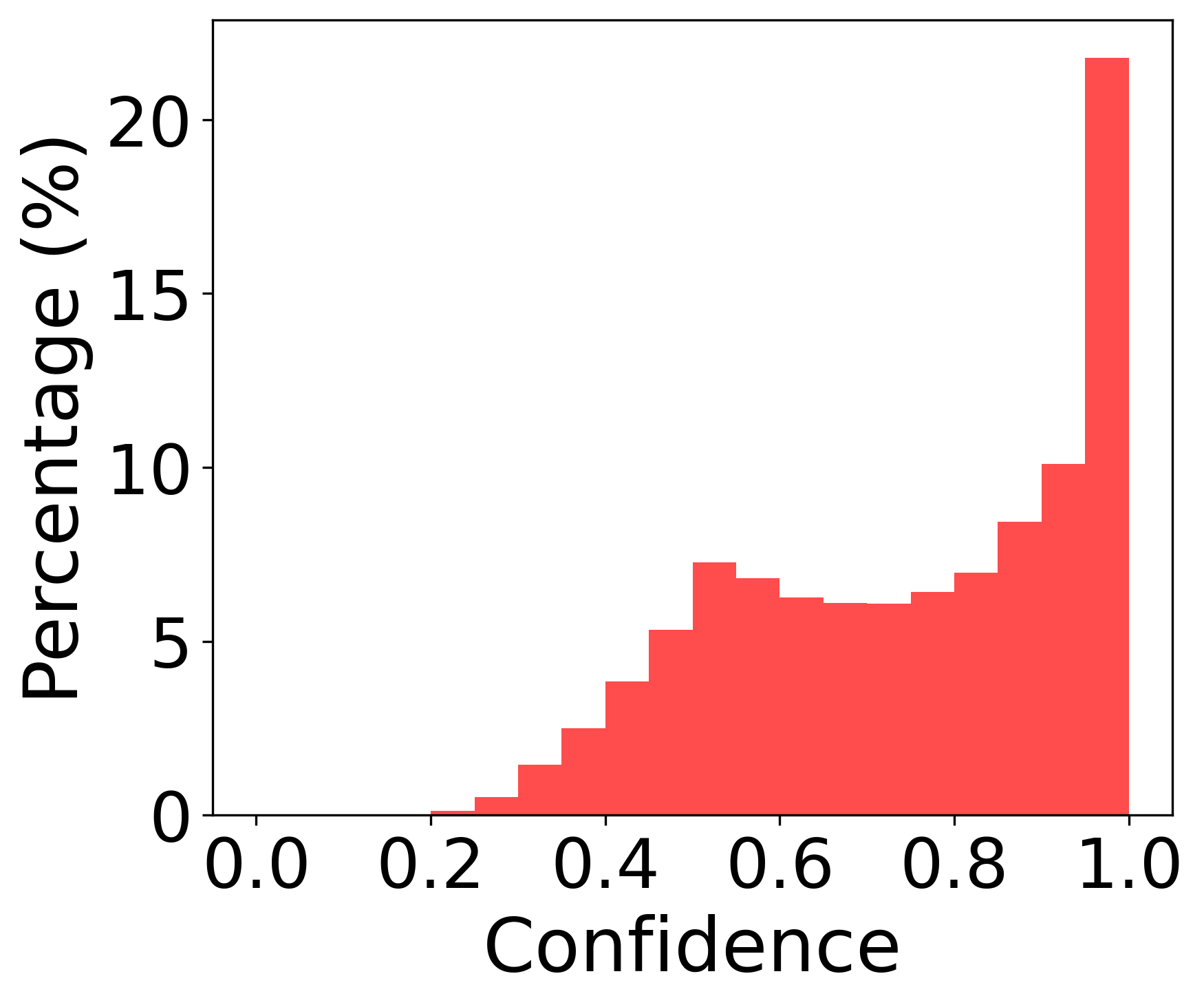}
\end{minipage}\hfill
\begin{minipage}[b]{0.245\linewidth}
  \centering
  \centerline{\tiny{PACS, ResNet-50}}
  \includegraphics[width=\linewidth]{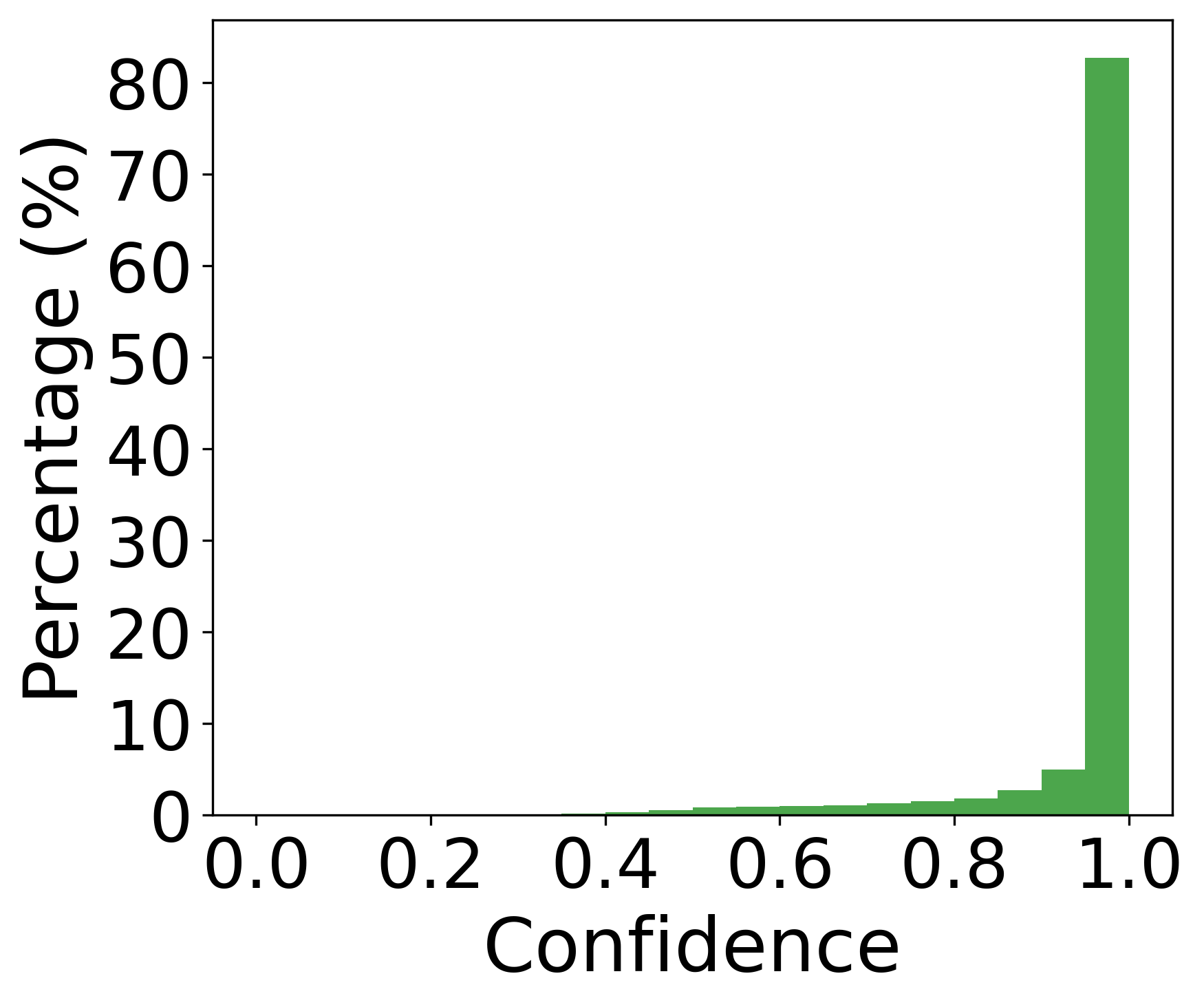}
\end{minipage}\hfill
\begin{minipage}[b]{0.245\linewidth}
  \centering
  \centerline{\tiny{PACS, ResNet-50}}
  \includegraphics[width=\linewidth]{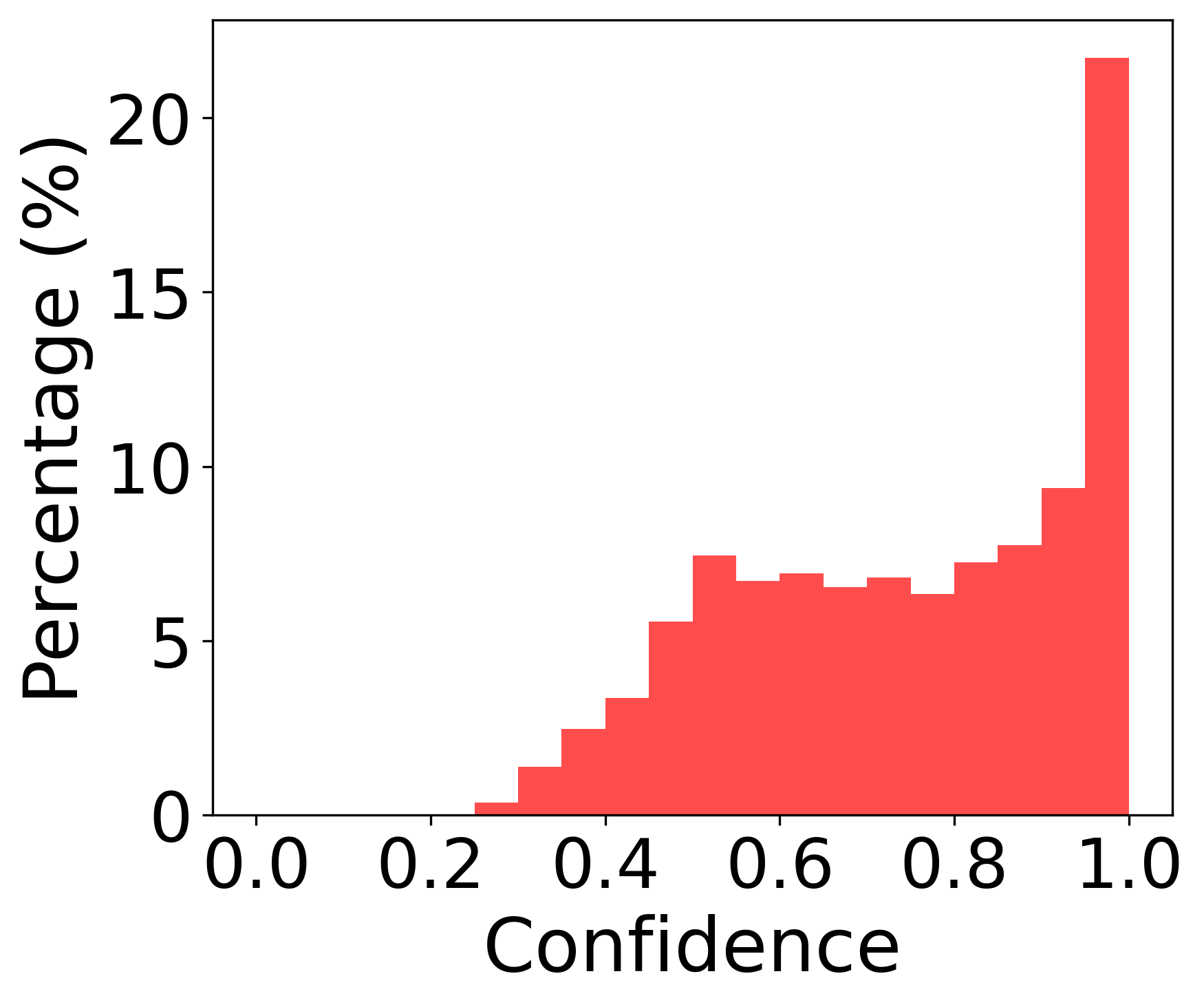}
\end{minipage}

\begin{minipage}[b]{0.245\linewidth}
  \centering
  \centerline{\tiny{Office-Home, ResNet-18}}
  \includegraphics[width=\linewidth]{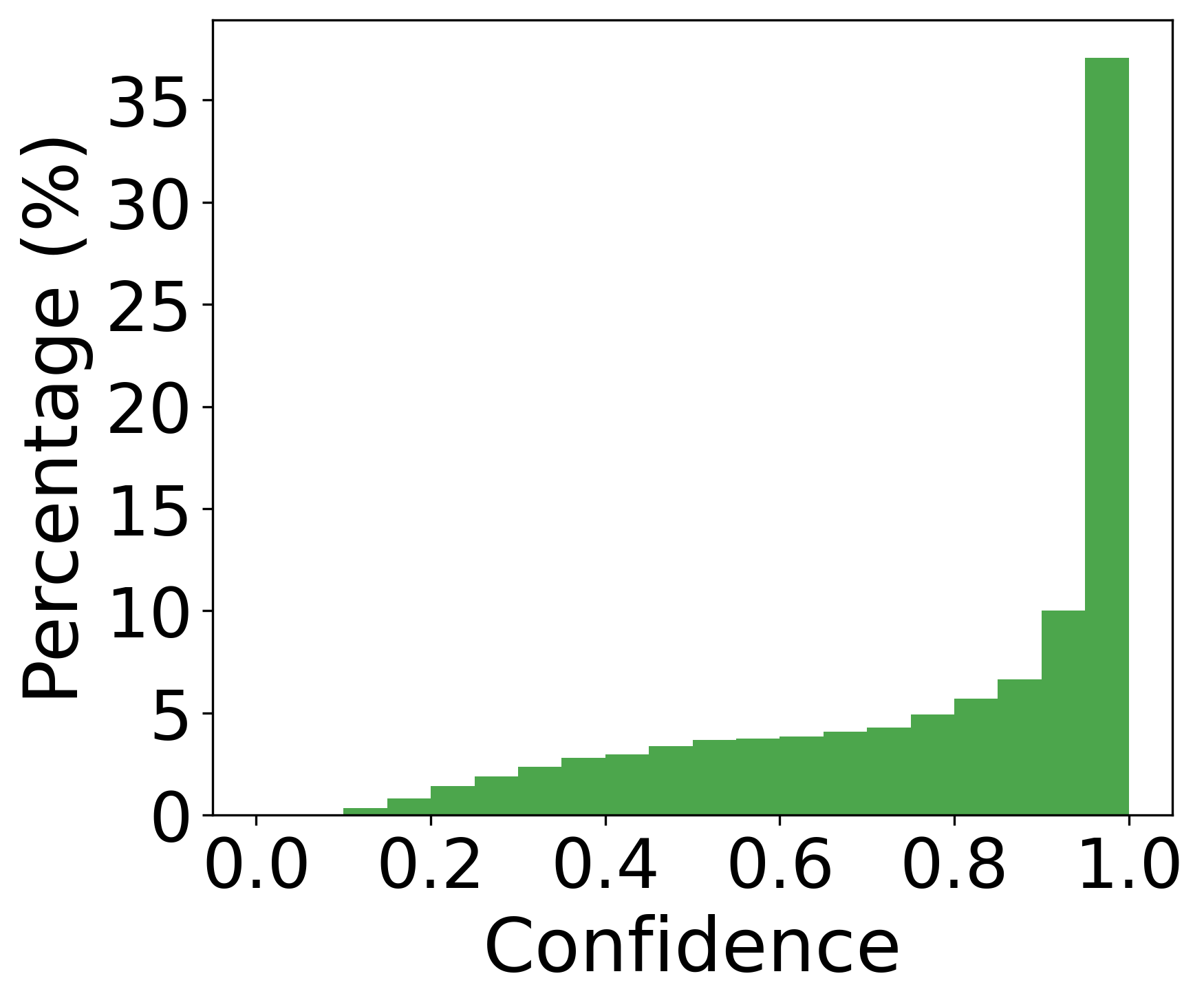}
\end{minipage}\hfill
\begin{minipage}[b]{0.245\linewidth}
  \centering
  \centerline{\tiny{Office-Home, ResNet-18}}
  \includegraphics[width=\linewidth]{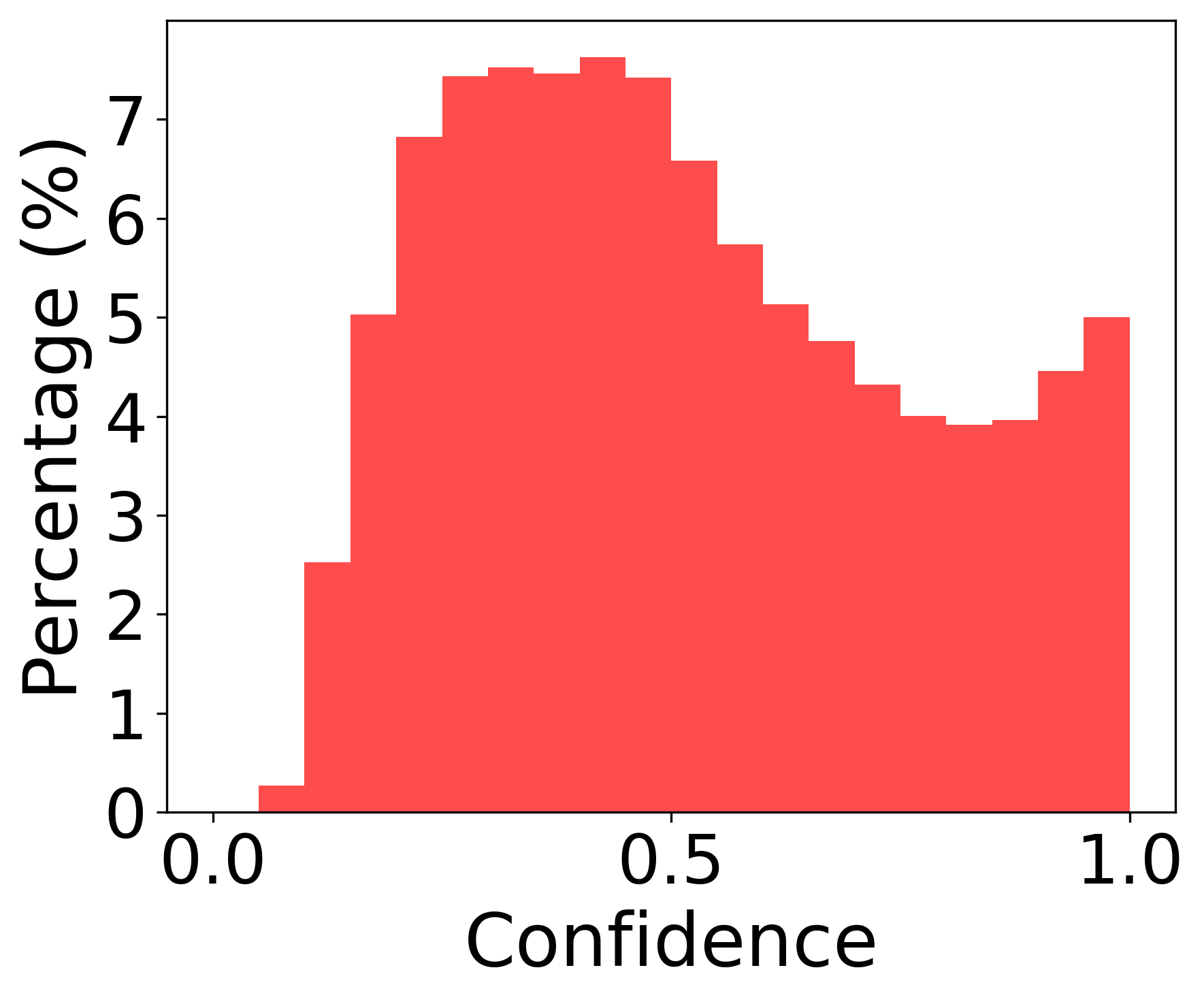}
\end{minipage}\hfill
\begin{minipage}[b]{0.245\linewidth}
  \centering
  \centerline{\tiny{Office-Home, ResNet-50}}
  \includegraphics[width=\linewidth]{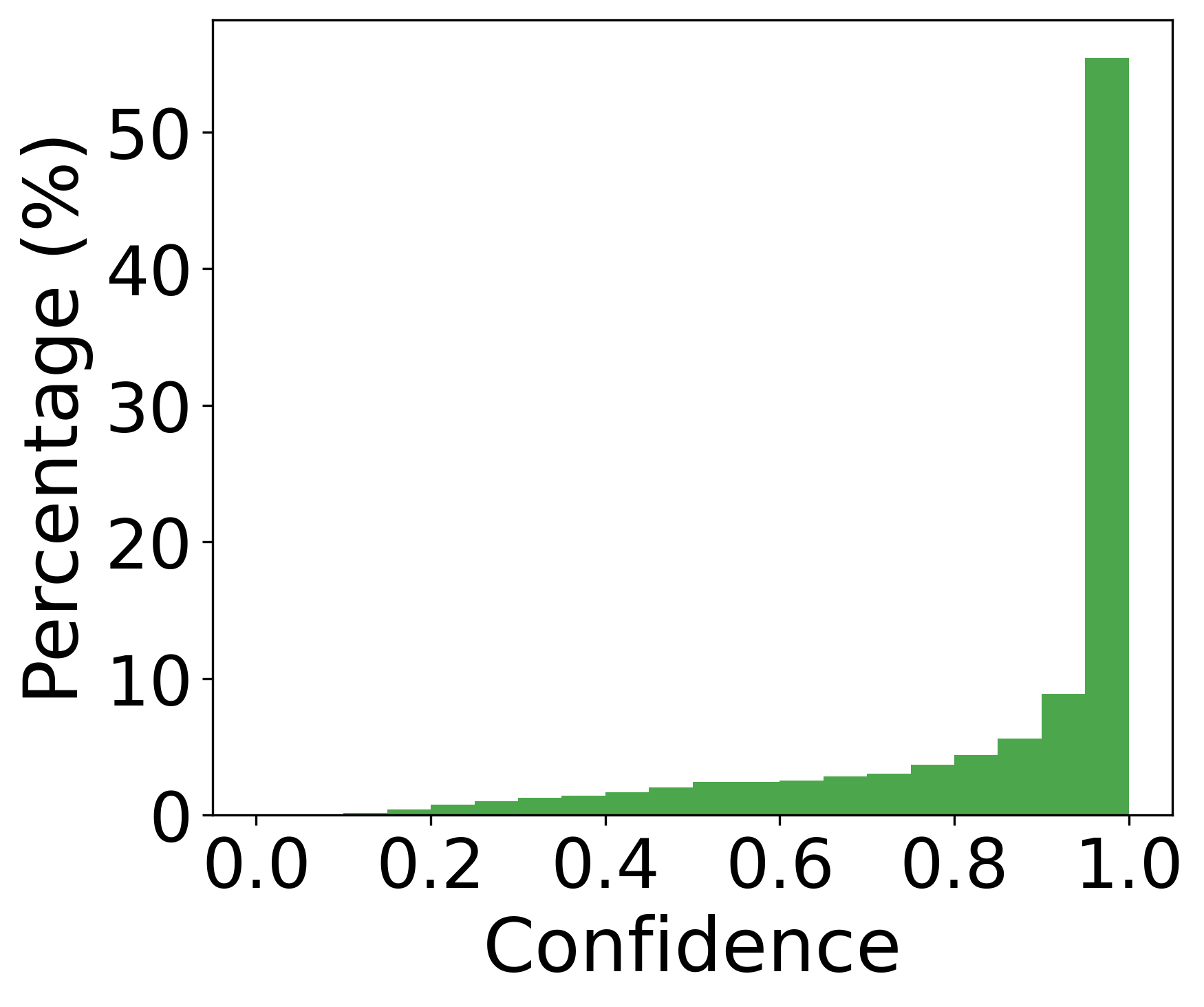}
\end{minipage}\hfill
\begin{minipage}[b]{0.245\linewidth}
  \centering
  \centerline{\tiny{Office-Home, ResNet-50}}
  \includegraphics[width=\linewidth]{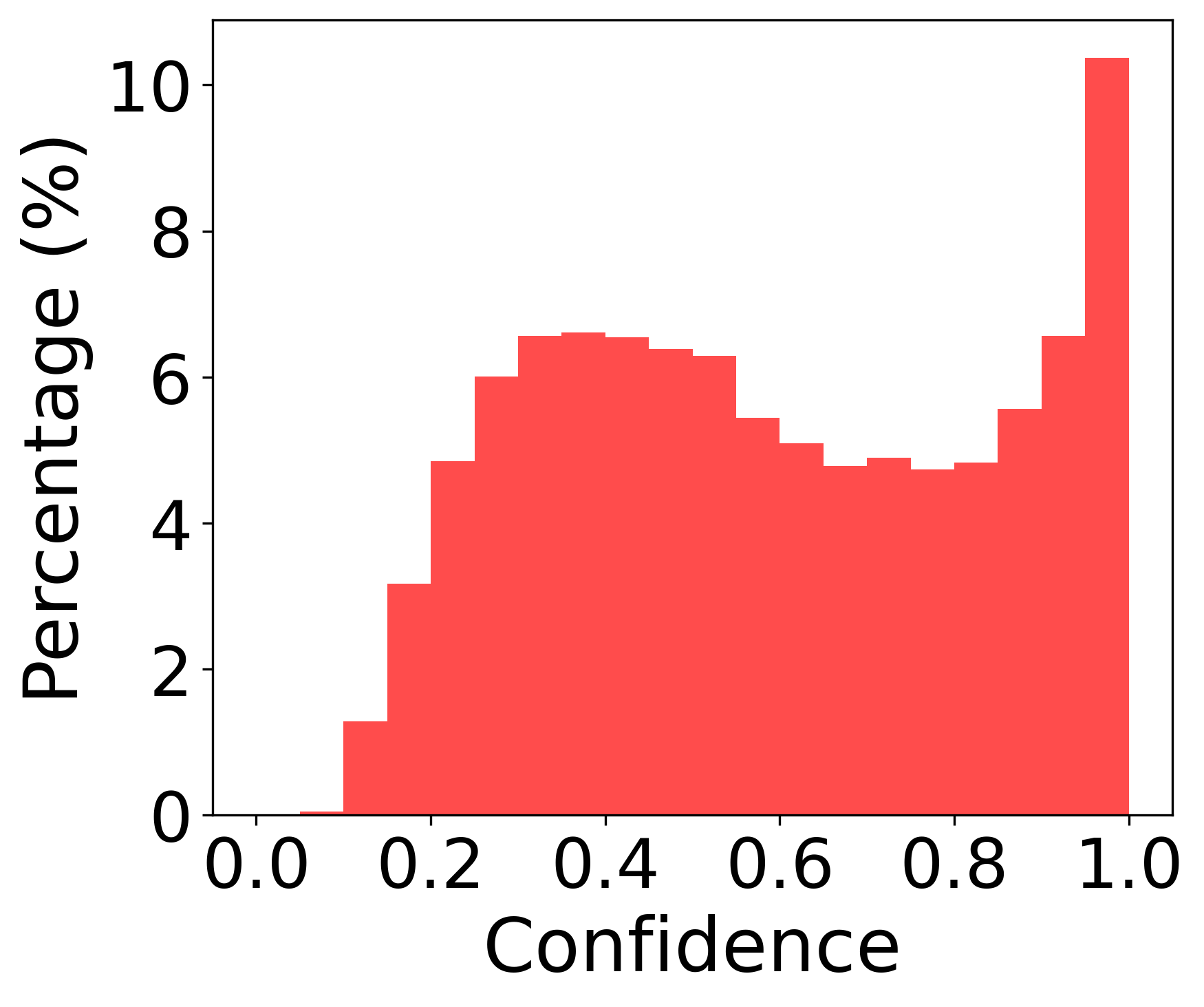}
\end{minipage}
\caption{Confidence distributions of CI-TTA before confidence filtering under ERM training. 
Top: PACS with ResNet-18/50; Bottom: Office-Home with ResNet-18/50. 
Green indicates correct predictions, while red indicates incorrect predictions.}
\label{fig:conf-dist}
\end{figure}

\textbf{Impact of Confidence Threshold}
We further investigate the effect of the filtering threshold $\tau$.  
Before reporting quantitative results, we examine the distribution of prediction confidence under deformation.  
Figure~\ref{fig:conf-dist} shows confidence histograms at $\sigma=0.01$ for CI-TTA before confidence filtering.
Across datasets and backbones, correct predictions are consistently skewed toward high confidence values (close to $1$), whereas incorrect predictions exhibit broader and left-shifted distributions.  
This clear separation suggests that thresholding can effectively suppress low-quality predictions while retaining the majority of correct ones.  
Such observations motivate the use of confidence-based filtering and the tuning of $\tau$ to balance error reduction against the risk of discarding true positives.  
To quantify this effect, we vary $\tau$ between $0.5$ and $0.9$ and report generalization accuracy on PACS in Table~\ref{tab:confidence_tta}.  
The results demonstrate that a moderate threshold ($\tau \in [0.6,0.7]$) achieves the most favorable balance, removing noisy low-confidence predictions while preserving the majority of reliable ones.  






\vspace{-1em}

\begin{table}[t]
    \centering
    \vspace{-1em}
    \caption{Effect of the confidence threshold $\tau$ in CI-TTA on the PACS dataset with ResNet-18 (ERM training). 
    Moderate thresholds ($\tau \in [0.6,0.7]$) provide the most favorable trade-off.}
    \label{tab:confidence_tta}
    \begin{tabular}{lcccccc}
        \hline
        \textbf{$\tau$} & \textbf{A} & \textbf{C} & \textbf{P} & \textbf{S} & \textbf{Avg} \\
        \cline{1-6}
        0.9 & 79.63 & 75.51 & 96.22 & 75.59 & 81.74 \\
        0.8 & 79.73 & 75.76 & 96.53 & 75.26 & 81.82 \\
        0.7 & \textbf{80.22} & 75.81 & \textbf{96.57} & 75.45 & \textbf{82.01}  \\
        0.6 & 79.98 & \textbf{75.85} & 96.40 & \textbf{75.48} & 81.92 \\
        0.5 & 79.88 & 75.81 & 96.39 & 75.46 & 81.88 \\
        \hline
    \end{tabular}   
\end{table}

\section{Conclusion}
\label{sec:conclusion}
In this work, we introduced Class-Invariant Test-Time Augmentation (CI-TTA), 
a lightweight inference-time strategy for improving domain generalization.  
By combining class-invariant deformations with a confidence-based filtering mechanism, 
CI-TTA promotes reliance on robust structural cues while simultaneously suppressing unreliable predictions.  
Extensive experiments on PACS and Office-Home, covering multiple DG algorithms and backbones, 
demonstrated consistent performance gains, confirming both the effectiveness and the generality of the approach.  
Despite its advantages, CI-TTA incurs additional inference cost and requires manual tuning of hyperparameters such as the deformation strength and confidence threshold.  
Future work will focus on developing adaptive mechanisms to automatically calibrate these parameters, and on exploring how CI-TTA can be integrated with training-time strategies to further enhance generalization.

\vfill\pagebreak

\clearpage
\bibliographystyle{IEEEbib}
\bibliography{strings}

\begin{thebibliography}{10}

\bibitem{ResNet}
Kaiming He, Xiangyu Zhang, Shaoqing Ren, and Jian Sun,
\newblock ``Deep residual learning for image recognition,''
\newblock in {\em Proceedings of the IEEE conference on computer vision and pattern recognition}, 2016, pp. 770--778.

\bibitem{wang2022generalizing}
Jindong Wang, Cuiling Lan, Chang Liu, Yidong Ouyang, Tao Qin, Wang Lu, Yiqiang Chen, Wenjun Zeng, and Philip~S Yu,
\newblock ``Generalizing to unseen domains: A survey on domain generalization,''
\newblock {\em IEEE transactions on knowledge and data engineering}, vol. 35, no. 8, pp. 8052--8072, 2022.

\bibitem{orhan2019robustness}
A~Emin Orhan,
\newblock ``Robustness properties of facebook's resnext wsl models,''
\newblock {\em arXiv preprint arXiv:1907.07640}, 2019.

\bibitem{ganin2016domain}
Yaroslav Ganin, Evgeniya Ustinova, Hana Ajakan, Pascal Germain, Hugo Larochelle, Fran{\c{c}}ois Laviolette, Mario March, and Victor Lempitsky,
\newblock ``Domain-adversarial training of neural networks,''
\newblock {\em Journal of machine learning research}, vol. 17, no. 59, pp. 1--35, 2016.

\bibitem{coral}
Baochen Sun and Kate Saenko,
\newblock ``Deep coral: Correlation alignment for deep domain adaptation,''
\newblock in {\em European conference on computer vision}. Springer, 2016, pp. 443--450.

\bibitem{mldg}
Da~Li, Yongxin Yang, Yi-Zhe Song, and Timothy Hospedales,
\newblock ``Learning to generalize: Meta-learning for domain generalization,''
\newblock in {\em Proceedings of the AAAI conference on artificial intelligence}, 2018, vol.~32.

\bibitem{vrex}
David Krueger, Ethan Caballero, Joern-Henrik Jacobsen, Amy Zhang, Jonathan Binas, Dinghuai Zhang, Remi Le~Priol, and Aaron Courville,
\newblock ``Out-of-distribution generalization via risk extrapolation (rex),''
\newblock in {\em International conference on machine learning}. PMLR, 2021, pp. 5815--5826.

\bibitem{groupdro}
Shiori Sagawa, Pang~Wei Koh, Tatsunori~B Hashimoto, and Percy Liang,
\newblock ``Distributionally robust neural networks for group shifts: On the importance of regularization for worst-case generalization,''
\newblock {\em arXiv preprint arXiv:1911.08731}, 2019.

\bibitem{rsc}
Zeyi Huang, Haohan Wang, Eric~P Xing, and Dong Huang,
\newblock ``Self-challenging improves cross-domain generalization,''
\newblock in {\em European conference on computer vision}. Springer, 2020, pp. 124--140.

\bibitem{zhang2017mixup}
Hongyi Zhang, Moustapha Cisse, Yann~N Dauphin, and David Lopez-Paz,
\newblock ``mixup: Beyond empirical risk minimization,''
\newblock {\em arXiv preprint arXiv:1710.09412}, 2017.

\bibitem{liang2025comprehensive}
Jian Liang, Ran He, and Tieniu Tan,
\newblock ``A comprehensive survey on test-time adaptation under distribution shifts,''
\newblock {\em International Journal of Computer Vision}, vol. 133, no. 1, pp. 31--64, 2025.

\bibitem{iwasawa2021test}
Yusuke Iwasawa and Yutaka Matsuo,
\newblock ``Test-time classifier adjustment module for model-agnostic domain generalization,''
\newblock {\em Advances in Neural Information Processing Systems}, vol. 34, pp. 2427--2440, 2021.

\bibitem{chen2023improved}
Liang Chen, Yong Zhang, Yibing Song, Ying Shan, and Lingqiao Liu,
\newblock ``Improved test-time adaptation for domain generalization,''
\newblock in {\em Proceedings of the IEEE/CVF Conference on Computer Vision and Pattern Recognition}, 2023, pp. 24172--24182.

\bibitem{zhang2022memo}
Marvin Zhang, Sergey Levine, and Chelsea Finn,
\newblock ``Memo: Test time robustness via adaptation and augmentation,''
\newblock {\em Advances in neural information processing systems}, vol. 35, pp. 38629--38642, 2022.

\bibitem{park2023test}
Jungwuk Park, Dong-Jun Han, Soyeong Kim, and Jaekyun Moon,
\newblock ``Test-time style shifting: Handling arbitrary styles in domain generalization,''
\newblock in {\em International Conference on Machine Learning}. PMLR, 2023, pp. 27114--27131.

\bibitem{zhao2022test}
Xingchen Zhao, Chang Liu, Anthony Sicilia, Seong~Jae Hwang, and Yun Fu,
\newblock ``Test-time fourier style calibration for domain generalization,''
\newblock {\em arXiv preprint arXiv:2205.06427}, 2022.

\bibitem{lyzhov2020greedy}
Alexander Lyzhov, Yuliya Molchanova, Arsenii Ashukha, Dmitry Molchanov, and Dmitry Vetrov,
\newblock ``Greedy policy search: A simple baseline for learnable test-time augmentation,''
\newblock in {\em Conference on uncertainty in artificial intelligence}. PMLR, 2020, pp. 1308--1317.

\bibitem{tta}
Divya Shanmugam, Davis Blalock, Guha Balakrishnan, and John Guttag,
\newblock ``Better aggregation in test-time augmentation,''
\newblock in {\em Proceedings of the IEEE/CVF international conference on computer vision}, 2021, pp. 1214--1223.

\bibitem{geirhos2018imagenetshape1}
Robert Geirhos, Patricia Rubisch, Claudio Michaelis, Matthias Bethge, Felix~A Wichmann, and Wieland Brendel,
\newblock ``Imagenet-trained cnns are biased towards texture; increasing shape bias improves accuracy and robustness,''
\newblock in {\em International conference on learning representations}, 2018.

\bibitem{ritter2017cognitiveshape2}
Samuel Ritter, David~GT Barrett, Adam Santoro, and Matt~M Botvinick,
\newblock ``Cognitive psychology for deep neural networks: A shape bias case study,''
\newblock in {\em International conference on machine learning}. PMLR, 2017, pp. 2940--2949.

\bibitem{castro2018elastic}
Eduardo Castro, Jaime~S Cardoso, and Jose~Costa Pereira,
\newblock ``Elastic deformations for data augmentation in breast cancer mass detection,''
\newblock in {\em 2018 IEEE EMBS International Conference on Biomedical \& Health Informatics (BHI)}. IEEE, 2018, pp. 230--234.

\bibitem{nguyen2020ensemble}
Tien~Thanh Nguyen, Anh~Vu Luong, Manh~Truong Dang, Alan Wee-Chung Liew, and John McCall,
\newblock ``Ensemble selection based on classifier prediction confidence,''
\newblock {\em Pattern Recognition}, vol. 100, pp. 107104, 2020.

\bibitem{wang2022self}
Xuezhi Wang, Jason Wei, Dale Schuurmans, Quoc Le, Ed~Chi, Sharan Narang, Aakanksha Chowdhery, and Denny Zhou,
\newblock ``Self-consistency improves chain of thought reasoning in language models,''
\newblock {\em arXiv preprint arXiv:2203.11171}, 2022.

\bibitem{pacs}
Da~Li, Yongxin Yang, Yi-Zhe Song, and Timothy~M Hospedales,
\newblock ``Deeper, broader and artier domain generalization,''
\newblock in {\em Proceedings of the IEEE international conference on computer vision}, 2017, pp. 5542--5550.

\bibitem{office-home}
Hemanth Venkateswara, Jose Eusebio, Shayok Chakraborty, and Sethuraman Panchanathan,
\newblock ``Deep hashing network for unsupervised domain adaptation,''
\newblock in {\em Proceedings of the IEEE conference on computer vision and pattern recognition}, 2017, pp. 5018--5027.

\bibitem{vapnik1998statistical}
Vladimir Vapnik and Vlamimir Vapnik,
\newblock ``Statistical learning theory wiley,''
\newblock {\em New York}, vol. 1, no. 624, pp. 2, 1998.

\end{thebibliography}

\end{document}